\documentclass{article}

    \PassOptionsToPackage{numbers, compress}{natbib}

\usepackage[preprint]{neurips_2022}

\usepackage[utf8]{inputenc} 
\usepackage[T1]{fontenc}    
\usepackage{hyperref}       
\usepackage{url}            
\usepackage{booktabs}       
\usepackage{amsfonts}       
\usepackage{nicefrac}       
\usepackage{microtype}      
\usepackage{xcolor}         
\usepackage{wrapfig}
\usepackage{graphicx}

\definecolor{babyblue}{rgb}{0.54, 0.81, 0.94}
\definecolor{asparagus}{rgb}{0.0, 0.5, 0.0}

\definecolor{ao(english)}{rgb}{0.0, 0.5, 0.0}
\title{Video ChatCaptioner:\\
Towards Enriched Spatiotemporal Descriptions}

\author{Jun Chen \qquad Deyao Zhu \qquad Kilichbek Haydarov \qquad Xiang Li  \qquad
 Mohamed Elhoseiny \\
King Abdullah University of Science and Technology\\
\small\texttt{\{jun.chen, deyao.zhu, kilichbek.haydarov, xiang.li.1, mohamed.elhoseiny\}@kaust.edu.sa}\\
}

\begin{document}
\maketitle

\begin{figure}[h!]
    \centering
    \includegraphics[width=0.85\linewidth]{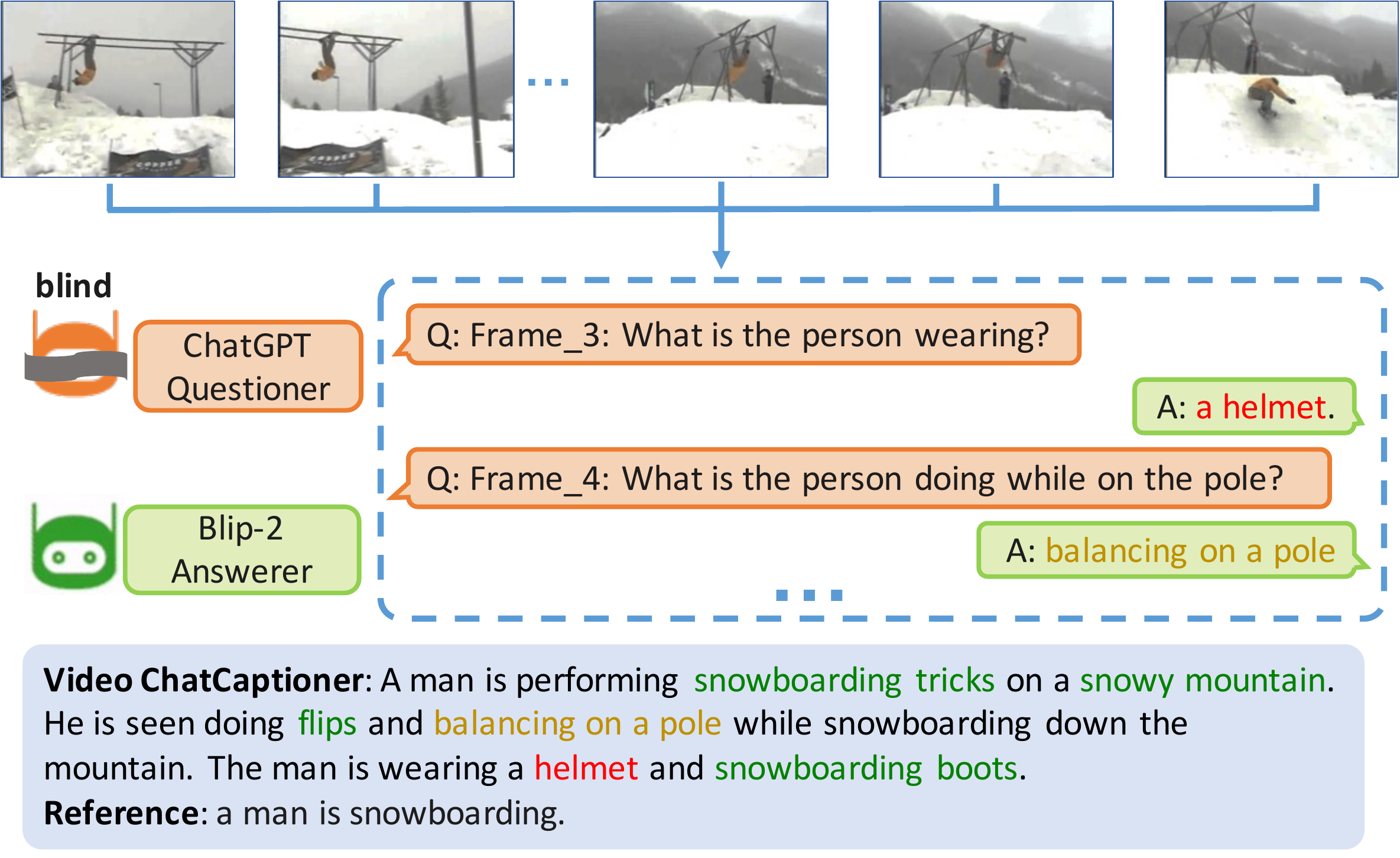}
    \caption{\textbf{Video ChatCaptioner} utilizes a ChatGPT to ask visual questions about a specific video, without access to the visual content. BLIP-2 serves as the answerer, providing answers to these questions based on the video content. Following multiple rounds of this interactive question-and-answer conversation, ChatGPT summarizes an enriched video description.}
    \label{fig_teaser}
\end{figure}

\begin{abstract}
Video captioning aims to convey dynamic scenes from videos using natural language, facilitating the understanding of spatiotemporal information within our environment. Although there have been recent advances, generating detailed and enriched video descriptions continues to be a substantial challenge. In this work, we introduce \emph{Video ChatCaptioner}, an approach for creating more comprehensive spatiotemporal video descriptions. Our method employs a ChatGPT model as a controller, specifically designed to select frames for posing video content-driven questions. Subsequently, BLIP-2 is utilized to answer these visual queries. This question-answer framework effectively uncovers intricate video details and shows promise as a method for enhancing video content. Following multiple conversational rounds, ChatGPT can summarize enriched video content based on previous conversations. 
Through the human evaluation experiments, we found that nearly 62.5\% of participants agree that Video ChatCaptioner can cover more visual information compared to ground-truth captions.
The code is publicly available at \url{https://github.com/Vision-CAIR/ChatCaptioner}



\end{abstract}

\section{Introduction}

Video captioning aims to automatically generate natural language descriptions by analyzing the visual information present in a given video. This technology offers numerous practical benefits, such as aiding the visually impaired~\cite{liu2021makes}, facilitating robotic navigation~\cite{huang2022visual}, and enhancing autopilot systems~\cite{rahman2019svin2}. Unlike image understanding, video captioning focuses on the spatiotemporal analysis of a dense sequence of frames that typically contain a wealth of visual information, including motion, people, and objects. Despite its potential, accurately capturing and representing the rich visual details within videos remains a formidable challenge for current prevailing approaches.

Enriched video captioning offers a more comprehensive understanding of video content for viewers. However, current methods face difficulties in generating detailed video captions due to two primary limitations. First, ground-truth captions in existing datasets tend to emphasize simple relationships or activities within a video, frequently neglecting more intricate information, such as more detailed descriptions of people, background elements, or specific actions. Second, the scale of video-caption datasets is considerably smaller compared to image-caption datasets. For example, the largest publicly available image-caption dataset, LAION-5B \cite{laion}, comprises 5 billion image-text pairs, while the largest video-text datasets, such as HowTo100M~\cite{howto100m} and WebViD-10M~\cite{frozen_in_time}, contain only 1.22 million and 10.7 million video-caption pairs, respectively. This disparity in scale poses a challenge for AI models to effectively learn and generate enriched and diverse video descriptions.

To tackle these challenges and produce more comprehensive video descriptions, we introduce Video ChatCaptioner. This method seeks to augment video understanding through a dialogue between the ChatGPT~\cite{chatgpt} and BLIP-2~\cite{blip2} models. While BLIP-2 is pretrained on hundreds of millions of image-text pairs, it struggles to process spatiotemporal data efficiently. The central innovation of Video ChatCaptioner involves utilizing ChatGPT as a controller, which then asks video content-driven questions to BLIP-2 for the sampled video frames to obtain a spatiotemporal understanding of the entire video. Finally, ChatGPT generates a video description based on the gathered question-answer pairs. Therefore, This approach holds the promise to eliminate the need for video-text datasets or models explicitly pretrained on such datasets, and also offers the potential to generate more enriched video captions. Additionally, we provide an illustrative example in Fig. \ref{fig_teaser}, depicting an example of our Video ChatCaptioner. It can be observed that Video ChatCaptioner produces more detailed descriptions in comparison to the ground truth. For examples, \textit{snowboarding tricks}, \textit{snowy mountain}, \textit{doing flips and balancing on a pole}, and \textit{helmet}.


We evaluated the performance of Video ChatCaptioner by generating captions for videos sampled from the MSVD~\cite{chen2011collecting} and WebVid~\cite{frozen_in_time} datasets. We conducted a qualitative comparison between the generated captions and the ground truth. The human evaluation experiments revealed that Video ChatCaptioner is capable of producing more comprehensive and informative video captions compared to ground-truth captions. Additionally, we demonstrated that our designed frameworks can generate a diverse set of visual questions, ensuring the extraction of more video content. Overall, Video ChatCaptioner presents a new paradigm for video caption generation, offering a valuable tool for crafting more intricate descriptions that effectively convey spatiotemporal information.



\section{Related work}

\noindent \textbf{Advancements in Image and Video Captioning.}
    Recent developments in pretraining models on extensive vision-language datasets have led to remarkable progress in both image~\cite{chen2020uniter,tsimpoukelli2021multimodal,kosmos,alayrac2022flamingo,visualgpt,blip1,blip2,florence,gpt4,beit3} and video~\cite{vid2seq,florence,frozen_in_time,vatt,fu2021violet,ge2022bridging,lei2021less,li2022align,li2020hero,miech2019howto100m,seo2022end,sun2019videobert,wang2022all,xue2022advancing} understanding. Significant advancements have been made in the field of image and video captioning, as demonstrated by various studies \cite{visualgpt, tsimpoukelli2021multimodal, alayrac2022flamingo, wang2022image, blip1, blip2, mplug2, vid2seq, florence, gpt4}. Upon pretraining on large-scale vision-language corpora, the resulting models are capable of generating more diverse and accurate descriptions for images or videos \cite{blip2,vid2seq,gpt4}. For instance, BLIP-2~\cite{blip2} and Flamingo~\cite{alayrac2022flamingo} exhibit remarkable image captioning performance after being pretrained on hundreds of millions of image-text pairs. Similarly, Vid2Seq~\cite{vid2seq} demonstrates improved temporal event localization and description accuracy after pretraining on an 18-million YT-Temporal-1B video dataset \cite{zellers2022merlot}. In more recent works, ChatCaptioner~\cite{chatcaptioner} explores the generation of more enriched image captions through interactions between ChatGPT~\cite{chatgpt} and BLIP-2, where ChatGPT serves for visual question asking, and BLIP-2 answers the questions. Our work is closely aligned with ChatCaptioner, with a focus on facilitating conversations between ChatGPT and BLIP-2 to produce richer video descriptions but more focusing on the spatiotemporal temporal understanding.


\noindent \textbf{Leveraging Pre-trained LLMs in Vision-Language Tasks.}
In recent years, research on employing autoregressive language models as decoders in vision-language tasks has gained significant momentum. This approach capitalizes on cross-modal transfer, enabling the transfer of knowledge between language and multimodal domains. VisualGPT~\cite{visualgpt} was among the first to demonstrate the data efficiency of utilizing a pre-trained language model, specifically GPT-2~\cite{gpt2}, in image captioning tasks. Building on this success, Frozen~\cite{tsimpoukelli2021multimodal} integrated a pre-trained vision encoder with GPT-2, further enhancing its performance. Subsequently, Flamingo~\cite{alayrac2022flamingo} introduced a gated cross-attention dense block to align the language model with a pre-trained vision encoder, while BLIP-2~\cite{blip2} proposed a Q-Former that transforms visual features into learnable visual tokens, making them more accessible to language models. KOSMAS-1 \cite{kosmos} further extended this approach by aligning various perception types (e.g., webpages, mathematical equations, and visual explanations) with a pre-trained LLM. Recently, after extensive pretraining, GPT-4~\cite{gpt4} has exhibited even more robust and strong visual understanding capabilities, emphasizing the continued advancements in this field.

\noindent \textbf{Interaction between LLMs and Other Modules.}
Large language models have demonstrated an impressive ability to improve their performance by integrating with external modules \cite{yao2022react, schick2023toolformer, visualChatGPT, chatcaptioner, mmreact, vipergpt, llm_augmenter}. For example, Toolformer \cite{schick2023toolformer} fine-tunes a 6.7B GPT-J model \cite{gpt-j} on a self-generated dataset with API-usage examples, enabling it to use external tools such as Wikipedia search API. ReAct \cite{yao2022react} prompts LLMs to generate reason traces explicitly, guiding the usage of external modules step-by-step for better performance. Visual ChatGPT \cite{visualChatGPT} and MM-ReAct \cite{mmreact} utilize ChatGPT's in-context learning ability to teach it how to use various vision models, enabling it to handle vision-related tasks.
Instead of using external modules with fixed API, ChatCaptioner \cite{chatcaptioner} employs another LLM model, BLIP-2 \cite{blip2}, as the external module, allowing ChatGPT to interact with it through natural language. This conversation between AI models results in better image captions with enriched details. Camel \cite{camel} and DERA \cite{nair2023dera} have further demonstrated the potential of conversational LLMs in programming and clinical tasks, respectively.
However, unlike ChatCaptioner, which is designed solely to describe images, Video ChatCaptioner leverages the temporal understanding ability of ChatGPT, empowering LLMs to generate rich and detailed descriptions of videos.


\section{Method}

\begin{figure*}
    \centering
    \includegraphics[width=1\linewidth]{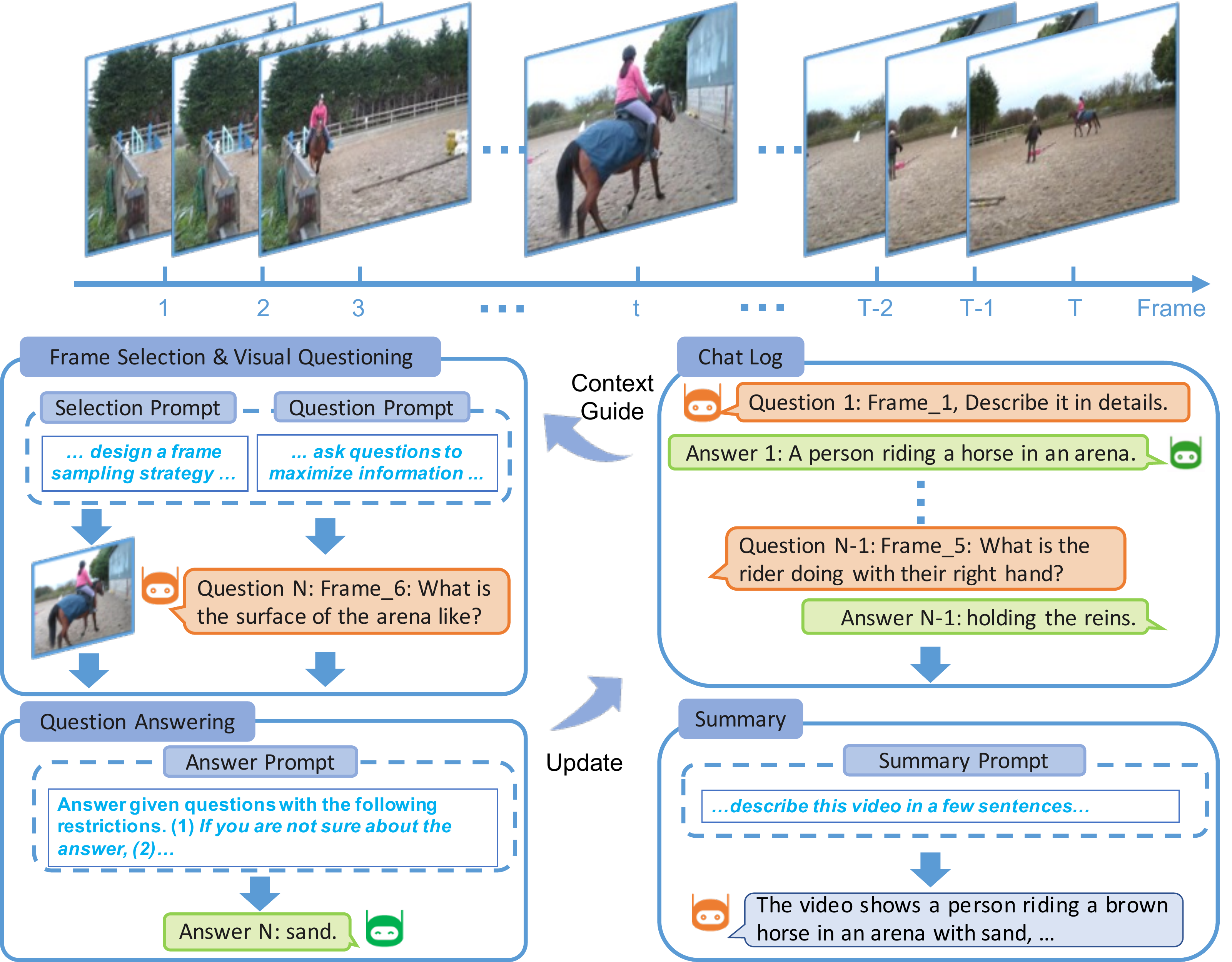}
    \caption{\textbf{ Video ChatCaptioner Pipeline:} Video ChatCaptioner incorporates two primary components: a questioner, ChatGPT, responsible for posing questions, and an answerer, BLIP-2, for answering the visual questions. With our designed instructions, ChatGPT is prompted to select a frame from a set of uniformly sampled frames and generate questions. Subsequently, BLIP-2 provides answers to the visual questions based on the chosen video frame. Finally, ChatGPT synthesizes a more enriched video caption from prior conversations. }
    \label{fig_pipeline}
\end{figure*}

Our Video ChatCaptioner is designed to capture the temporal relationships among distinct spatial features by leveraging the automatic questioning and summarization capabilities of ChatGPT. To achieve this, ChatGPT poses a variety of questions to BLIP-2, a vision-language model proficient in answering diverse visual queries. Although BLIP-2 is only trained on image-text pairs, it can effectively infer various motion information, such as dancing, sweeping, and riding, by only referencing individual frames. However, a single BLIP-2 model cannot adequately represent the complete temporal information within a video. Therefore, we propose Video ChatCaptioner, in Fig. \ref{fig_pipeline}, which treats ChatGPT as a controller to ask visual questions across different frames and asks BLIP-2, as a VQA model, to provide the answers. After multiple conversations between ChatGPT and BLIP-2, ChatGPT will aggregate the spatiotemporal information together to summarize the video content.

\begin{figure*}
    \centering
    \includegraphics[width=0.9\linewidth]{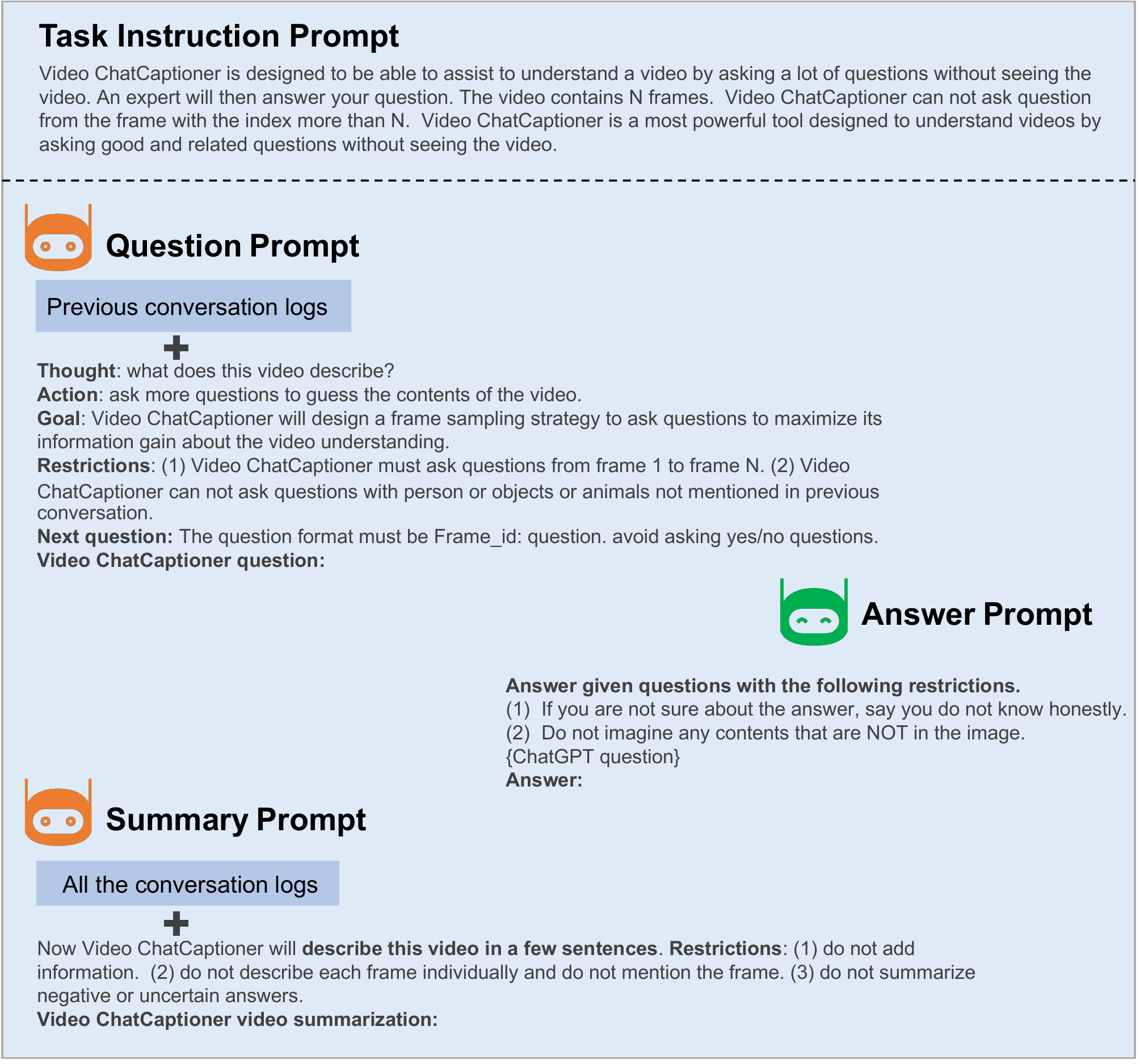}
    \caption{\textbf{Prompt design}: We showcase the prompt system designed for our Video ChatCaptioner, which incorporates task instruction, question, answer and summary prompts to facilitate effective communication between ChatGPT and BLIP-2.}
    \label{prompt_design}
\end{figure*}

In order to achieve our goal, we initially uniform sample $N$ frames from a given video, and we ask ChatGPT to ask informative questions for each frame. Since ChatGPT does not have visual perception ability, we ask BLIP-2 to answer the visual questions from ChatGPT. We have developed a prompt system to guide ChatGPT in generating a diverse set of informative questions. We demonstrate our designed prompt in Fig \ref{prompt_design}. Our prompt system comprises several levels, including task instruction, sub-question instruction, and summary instruction, each designed to enhance the quality of the questions generated. Additionally, we have designed prompts to encourage BLIP-2 to generate more reliable answers. In the following section, we will describe each level of the prompt system in details.

\subsection{Prompt Design}
\noindent \textbf{Task instruction for video understanding.} The purpose of task instructions is to offer a comprehensive context and outline the specific task for ChatGPT in the domain of video content understanding. By doing so, it effectively steers ChatGPT towards generating more targeted, video-content-related questions. The effective communication between ChatGPT and BLIP-2 heavily depends on the proper design of these instructions. They allow the information about the video content to be enriched during each conversation round, leading to deeper and more engaging interactions between the two agents.


\noindent \textbf{Incorporating conversation logs.} To enhance the quality of questions generated by ChatGPT, we incorporate the conversation history prior to posing a new question. This is achieved by pre-appending a series of prior question-answer pairs to the question prompt. Supplying ChatGPT with this accumulated contextual information can encourage the model to generate more insightful and relevant questions.

\noindent \textbf{Question prompt.} The sub-question prompt aims to leverage the history of question-answer pairs to generate more informative and relevant questions while adhering to specific constraints. By reasoning from the available question-answer pairs, the prompt provides explicit restrictions to guide the generation of these questions, ensuring that they are coherent, relevant, and aligned with the objectives. We demonstrate the sub-questions prompt as follows. 






For the first question, we use a fixed prompt to request a detailed description from the first frame, as shown below:

\textit{\quad Frame\_1: Describe it in details.}

For subsequent questions, we guide ChatGPT to formulate inquiries by defining clear objectives, specifying the necessary actions, and establishing constraints for the next question.

\textbf{\quad -  Frame sampling design:} Rather than employing random frame sampling, we encourage ChatGPT to devise a best frame sampling strategy on its own under our designed instructions. This approach allows the model to select the frames based on the current context to maximize its understanding of the whole video content. 

\textbf{\quad -  Question restriction:} We impose specific limitations on the generation of follow-up questions. (1) These questions must adhere to the format of $Frame\_ID$, enabling us to easily determine the frame ID with regular expression for visual questioning. We also confine the frame selection range to fall between 1 and N, the maximum sampled frame number, which can prevent ChatGPT from selecting frames that are out-of-bounds. (2) Moreover, we prohibit Video ChatCaptioner from posing questions related to unmentioned individuals or objects from prior conversations. This measure aims to reduce the occurrence of hallucinations during the visual question answering stage. (3) Furthermore, to encourage the formulation of more informative inquiries, we restrict ChatGPT from generating yes/no questions.

\begin{figure}
    \centering
    \includegraphics[width=1\linewidth]{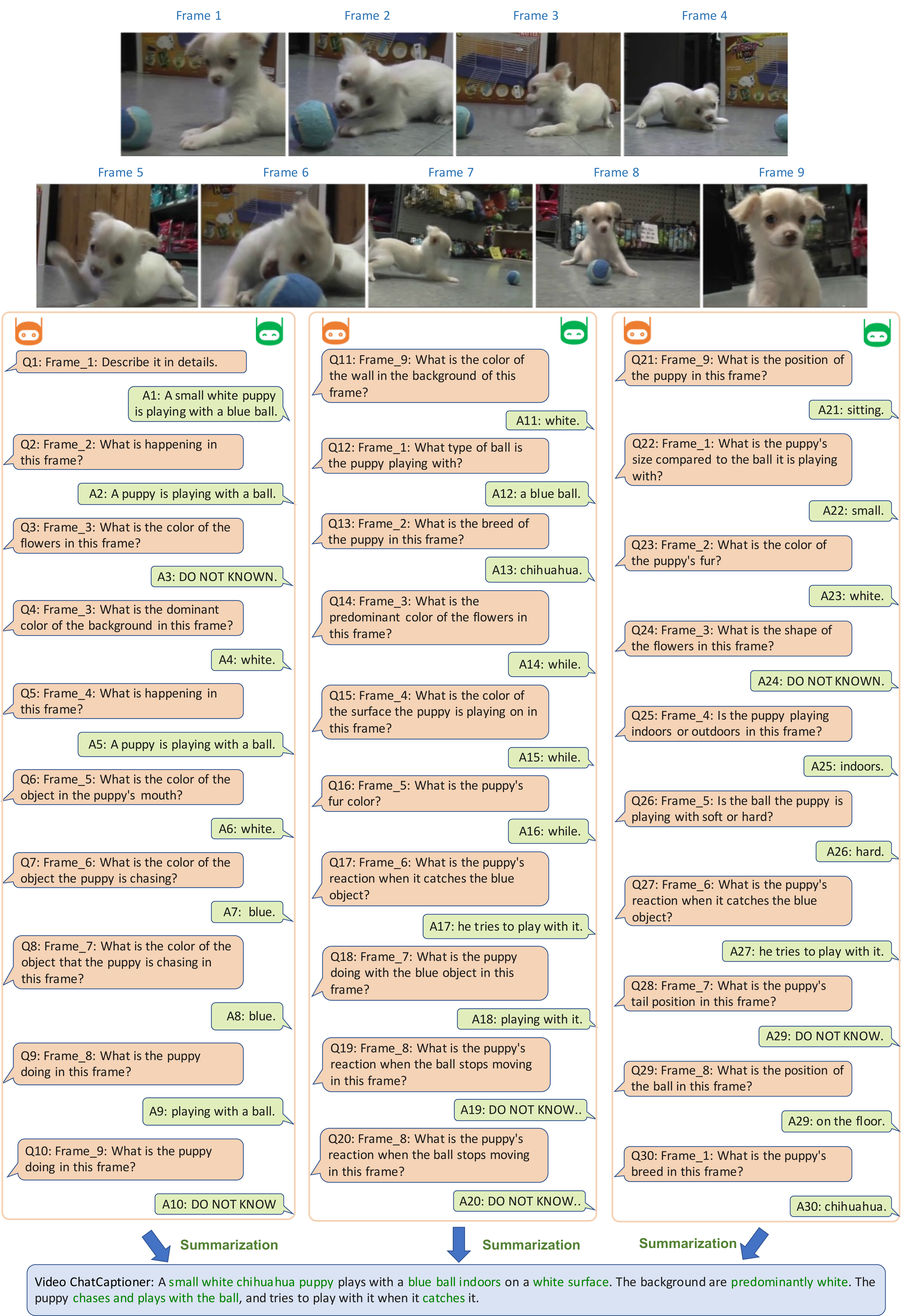}
    \caption{Illustrating ChatGPT and BLIP-2 conversation: We uniformly sample 9 frames from an MSVD video~\cite{chen2011collecting} and prompt ChatGPT to autonomously select frames and generate 30 distinct questions for visual questioning }
    \label{fig_conversation}
\end{figure}

\noindent \textbf{BLIP-2 answer prompt.} We employ BLIP-2 to supply answers to ChatGPT's questions. As ChatGPT may pose highly open-ended and complex questions, BLIP-2 could potentially generate incorrect answers due to its language inductive bias. To mitigate this issue, we have tailored the prompt to minimize such occurrences. We allow BLIP-2 to respond with "do not know" if it is uncertain about an answer, which has been empirically shown to reduce the number of unconfident responses effectively.

\noindent \textbf{Summary Prompt.} In order to create a thorough and accurate summary of the previous conversation between ChatGPT and BLIP2, we have carefully crafted a prompt that encourages ChatGPT to synthesize the most relevant and precise information from the prior question-answer exchanges. Examples of such prompts can be found in Fig. \ref{prompt_design}.






\section{Experiment}

In this section, we will describe our Video ChatCaptioner from multiple perspectives. First, we will discuss the details of the experimental setting, the primary experimental results, and also ablation studies for different prompt designs.

\textbf{Experimental Setup.} Our experiment primarily involves ChatGPT~\cite{chatgpt} and BLIP-2~\cite{blip2} engaging in conversation with each other to obtain the spatiotemporal perception of a given video. We specifically utilize the "gpt-3.5-turbo" version of ChatGPT, accessible via the OpenAI API~\cite{chatgptapi}. Additionally, we employ the BLIP-2 version, which incorporates a FlanT5-XXL~\cite{flanT5} language model and a ViT-G vision encoder from the EVA-CLIP framework~\cite{fang2022eva}. We sample the images from WebViD~\cite{frozen_in_time} and MSVD~\cite{chen2011collecting} datasets.

\begin{wraptable}{r}{0.45\textwidth}
\centering
\scalebox{0.9}{
\begin{tabular}{ccccc}
\toprule
Vote percentage & WebViD & MSVD & Avg \\
\midrule
GT & 40\% & 35\% & 37.5\% \\
Ours & \textbf{60\%}  & \textbf{65\%} & \textbf{62.5\%} \\
\bottomrule
\end{tabular}
}
\caption{Human votes between ground-truth video caption and the captions generated by our Video Chatcaptioner.}
\vspace{-1em}
\label{human_evaluation}
\end{wraptable}


\begin{figure}
    \centering
    \includegraphics[width=1\linewidth]{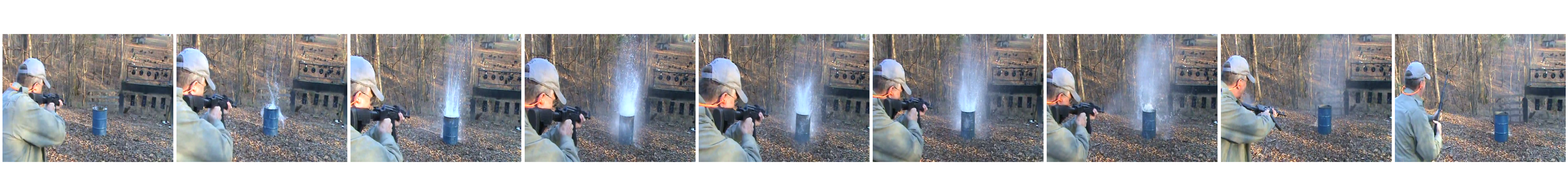}
    \begin{flushleft} 
    \textbf{Reference:}{\fontfamily{qtm}\selectfont  a man is rapidly firing bullets at a drum filled with water.} \\ \vspace{3pt} \textbf{Frame-centric sampling:} {\fontfamily{qtm}\selectfont The video shows a man in a \textcolor{asparagus}{wooded area} wearing a \textcolor{asparagus}{hat} and \textcolor{asparagus}{shooting a rifle} at various targets, including \textcolor{asparagus}{a white target, a red target, a water spout, and a barrel}. He is seen \textcolor{asparagus}{standing, crouching}, and aiming with a \textcolor{asparagus}{serious expression} on his face. The man is also shown \textcolor{asparagus}{wearing a jacket} in some frames. The video captures the motion and activity of the man as he \textcolor{asparagus}{shoots the gun} and \textcolor{asparagus}{moves around} the forest.} \\ \vspace{3pt}
    \textbf{Video-centric sampling:} {\fontfamily{qtm}\selectfont A man is \textcolor{asparagus}{shooting a pistol} in a \textcolor{asparagus}{wooded area}. He is wearing a \textcolor{asparagus}{hat} and shooting at a \textcolor{asparagus}{stationary object} with \textcolor{asparagus}{multiple shots} at once. He is \textcolor{asparagus}{shooting rapidly with two hands} and \textcolor{asparagus}{towards the ground}.}
    \end{flushleft}
    \caption{frame-centric vs. video-centric sampling. The key objects and actions captured are highlighted in green color.}
    \vspace{-10pt}
    \label{sampling_abltion}
\end{figure}

\noindent \textbf{Human evaluation:} In order to assess the effectiveness of our Video ChatCaptioner in generating more informative video captions, we conducted a random sampling of 100 images each from the MSVD~\cite{chen2011collecting} and WebViD~\cite{frozen_in_time} datasets. We then compared the generated captions  with the ground-truth (GT) captions using Amazon Mechanical Turkers as evaluators. For each video, five human participants were engaged in the evaluation process, comparing both the GT caption and our generated caption. Participants were asked to determine which caption covers more accurate video information, such as objects and visual relationships. Caption preference was determined based on majority vote. 

The results, as displayed in Table \ref{human_evaluation}, reveal that the Video ChatCaptioner received a preference of 62.5\% compared to the 37.5\% for the ground-truth captions. This outcome suggests that the Video ChatCaptioner is capable of generating more comprehensive and detailed video captions than the ground-truth alternatives.

\noindent \textbf{Full conversation demonstration:} We randomly sampled an image from the MSVD dataset~\cite{chen2011collecting} and presented the complete conversation along with the resulting summarized video description in Fig. \ref{fig_conversation}. The full dialog between ChatGPT and BLIP-2 shows various aspects, including the selection of frames by ChatGPT for posing questions, BLIP-2's responses to those questions, and BLIP-2's handling of uncertain queries, among other interactions. The Video ChatCaptioner ultimately presents a comprehensive video summary. This summary effectively identifies several crucial elements, such as a \textit{small white chihuahua puppy},  a \textit{white surface}, and the \textit{chasing and playing with the ball} behaviors.



\noindent \textbf{Frame sampling strategy ablation.} There are two primary approaches guiding ChatGPT's question-asking process. The first referrs to as the frame-centric approach, where ChatGPT asks questions from each frame individually and generates a complete caption for each frame. Finally, ChatGPT synthesizes a video summary based on these individual frame captions.  The second, the video-centric approach, requires ChatGPT to function within a broader video context.  ChatGPT autonomously selects the frame indices for asking questions and, after a series of queries, generates a video summary based on the prior conversation. 

To evaluate the effectiveness of these distinct methodologies, we conducted an ablation study comparing the frame-centric and video-centric strategies using the MSVD videos~\cite{chen2011collecting}. A qualitative example from this analysis is illustrated in Fig \ref{sampling_abltion}. The results reveal that the frame-centric sampling strategy tends to generate more inaccurate descriptions. For instance, it incorrectly identifies a single shooting target as multiple entities, such as a white target, a red target, a water spout, and a barrel. In contrast, the video-centric strategy accurately recognizes the object as a stationary target. This discrepancy arises because the frame-centric sampling approach fails to preserve the temporal consistency of objects within the video.

\begin{figure}
    \centering
        \includegraphics[width=1\linewidth]{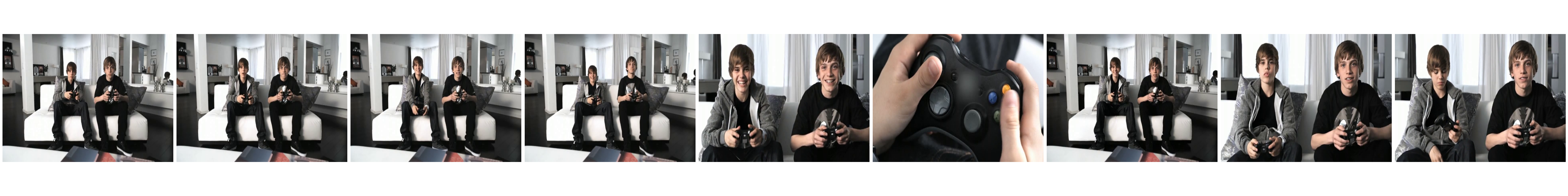} \\
    \vspace{-10pt}
    \begin{flushleft}
    \textbf{Reference:} {\fontfamily{qtm}\selectfont two boys playing video game.} \\
    \textbf{Video ChatCaptioner:} {\fontfamily{qtm}\selectfont Two boys are \textcolor{asparagus}{sitting on a white couch} in a living room \textcolor{asparagus}{playing video games}. They are wearing \textcolor{asparagus}{black hoodies} and playing with \textcolor{asparagus}{an Xbox 360 controller} connected to a TV. A laptop is visible on the \textcolor{asparagus}{table} in front of them. The boys seem to be \textcolor{asparagus}{enjoying the game.}}\\
    \end{flushleft}
    \includegraphics[width=1\linewidth]{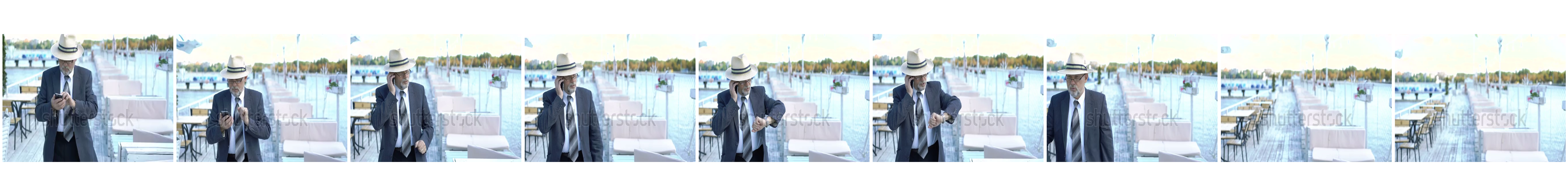} 
    \vspace{-15pt}
    \begin{flushleft} 
    \textbf{Reference:} {\fontfamily{qtm}\selectfont Elderly businessman calling on phone on the open cafe near sea.\\ \textbf{Video ChatCaptioner:} A man in a \textcolor{asparagus}{suit and hat} is \textcolor{asparagus}{standing on a dock looking at his phone} and \textcolor{asparagus}{talking} on it. He appears to be \textcolor{asparagus}{checking the time on his watch} as well. In the background, \textcolor{asparagus}{a city} can be seen.} 
    \end{flushleft}
    \caption{More sampled examples with our Video ChatCaptioner. The key objects and actions are highlighted in green color. }
    \label{more_examplesss7}
    \vspace{-15pt}
\end{figure}

\begin{figure}
    \centering
    \includegraphics[width=0.8\linewidth]{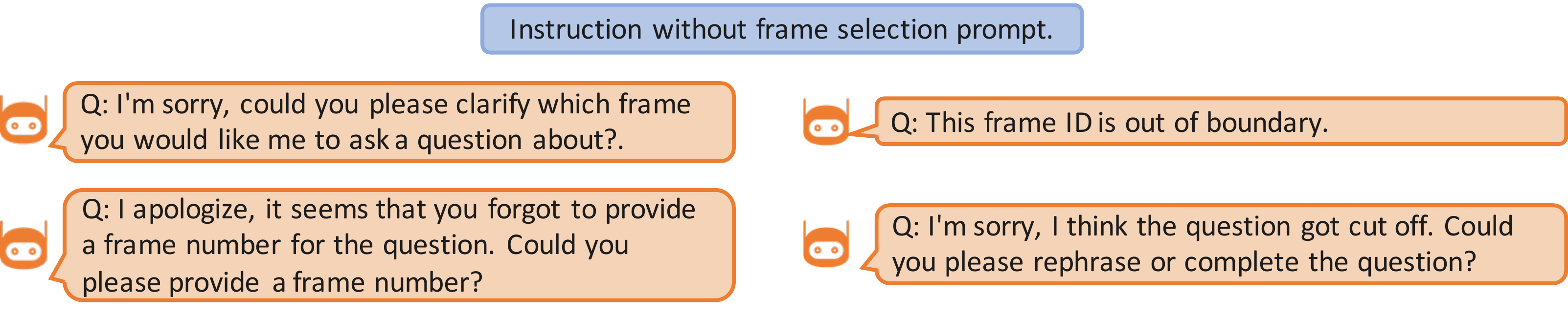}
    \caption{The effect of without applying the frame selection prompt on the question instruction.}
    \label{fig_frame_sel}
     \vspace{-15pt}
\end{figure}


\begin{wraptable}{r}{0.45\textwidth}
  \begin{center}
\scalebox{0.9}{
\begin{tabular}{cc}
\toprule
Per Dialogue & All Questions \\
\midrule
26.5/29 & 3326/4350 \\
91\% & 76\% \\
\bottomrule
\end{tabular}
}
  \end{center}
  \caption{Number of unique questions per dialogue and per all questions.}
\label{unique_question}
\vspace{-10pt}
\end{wraptable}
\noindent \textbf{Question Diversity:} 
We evaluate the diversity of generated questions by couting the uniqueness of teh questions posed during dialogues.
We randomly select 150 images from the MSVD dataset and generate 30 questions for each conversation with Video ChatCaptioner. Given that the initial question is always hard-coded, we exclude it from our analysis. The results, presented in Table \ref{unique_question}, reveal that 91\% of the questions within each dialogue and 76\% of the questions across the entire dataset are unique. This substantial proportion of unique questions underscores the diversity of the questions generated by Video ChatCaptioner.


    



\noindent \textbf{Additional Qualitative Examples:} In Fig \ref{more_examplesss7}, we present further qualitative instances that showcase the superior performance of Video ChatCaptioner in generating detailed and informative video description compared to the ground truth. 
For instance, in the first example, Video ChatCaptioner accurately identifies that two girls are performing a synchronized choreographed routine, dressed in pink and different short dresses and dancing barefoot. These intricate details offer a more vivid and comprehensive description of the video, which is absent in the ground truth captions.

\noindent \textbf{Optimizing frame ID selection in ChatGPT.} ChatGPT may occasionally fail to generate the desired Frame\_ID format, such as producing an out-of-bounds frame index or not adhering to the required format. To ensure that ChatGPT generates the appropriate frame ID format, we introduce two prompts in the questions prompt:
\textit{
(1) Video ChatCaptioner must ask questions from frame 1 to frame N. (2) The questions format must be Frame\_ID: question.}
Subsequently, we employ a regular expression to extract the frame ID. A qualitative ablation study, presented in Fig. \ref{fig_frame_sel}, demonstrates the effectiveness of these prompts in refining frame ID selection.


\section{Limitation}

Through the success of Video ChatCaptioner capable of generating more diverse and detailed video descrition, the video ChatCaptioner system still suffers from several limitations.

\textbf{Limitations of the perception model:} While BLIP-2 exhibits robust visual perception and VQA capabilities, it occasionally misidentifies objects, attributes, or visual relationships, which can compromise the accuracy of the generated video captions. Moreover, as BLIP-2 is not specifically trained on videos, it may struggle to capture intricate motion information. In addition, due to its training on a static large language model, BLIP-2 inadvertently inherits language biases present in the LLM, potentially generating content that is not visually grounded. Addressing these limitations can be accomplished by employing a more advanced perception model, such as GPT-4~\cite{gpt4}.

\textbf{Inference Speed:} The Video ChatCaptioner system necessitates the collaboration of a language model and a VQA (Visual Question Answering) model to generate descriptive video captions through interactive conversations. As a result of these multi-round conversations, the inference time for the Video ChatCaptioner is typically longer compared to traditional video captioning models. To optimize the inference speed, one possible solution is to instruct the language model to formulate more informative questions so that the number of conversation rounds can be minimized.

\begin{figure}
    \centering
    \includegraphics[width=1\linewidth]{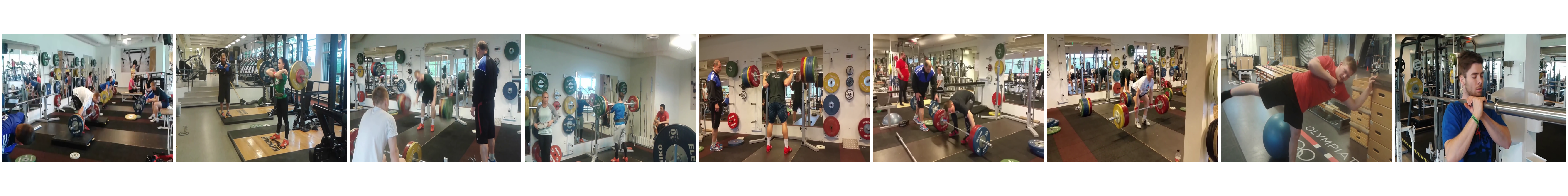} \\
    \vspace{-15pt}
    \begin{flushleft}
    \textbf{Reference:} A large group of people are seen sitting around a gym with many seen on the side lifting weights over their heads. \\ \textbf{Ours:} In the video, a man is seen lifting a barbell and doing squats in a gym. The man is wearing a shirt and shorts, and is not using any other equipment besides the barbell. A woman is also seen lifting weights in the background.  The gym also has a bench. \\
    \end{flushleft}
    \caption{Illustrated example that fails to temporally capture the multiple people}
    \label{fig_bad}
    \vspace{-20pt}
\end{figure}

\textbf{Temporal Identification of multiple people or objects:} In Video ChatCaptioner, ChatGPT remains blind to the actual video content, hence it lacks an understanding of temporal consistency information. For instance, Fig. \ref{fig_bad} illustrates that a video depicts a group of people lifting barbells in a gym, appearing in different frames. However, the Video ChatCaptioner only mentions a man and a woman. This issue may arise from two primary factors: 1) The uniform sampling of video frames fails to capture objects in videos with rapidly changing frame rates, which hinders BLIP-2's ability to effectively process multiple objects or people information; 2) ChatGPT can only identify objects or people from text conversations, lacking visual grounding capabilities, which may result in insufficient coverage of multiple objects or people in its output.

\section{Discussion}

In this work, we introduce a novel approach, dubbed Video ChatCaptioner, which generates enriched spatiotemporal descriptions through interactive conversations between ChatGPT and BLIP-2. ChatGPT serves as a controller, selecting frames for posing visual questions, while BLIP-2 provides answers to these questions. Subsequently, ChatGPT synthesizes a comprehensive video description by integrating information from the preceding dialogue. Through qualitative examples and also the human evaluation experiments, we observe that our method can generate richer video descriptions compared to ground-truth captions. Overall, our approach provides a new paradigm for video caption generation and we hope to inspire further research in this area.






{\small
\bibliographystyle{ieee_fullname}
\bibliography{egbib}
}

\end{document}